\documentclass{sig-alternate-2013}
\usepackage{hyperref}
\usepackage{array,multirow,graphicx}
\usepackage[table]{xcolor}
\usepackage{threeparttable}
\usepackage{color,soul}

\newfont{\mycrnotice}{ptmr8t at 7pt}
\newfont{\myconfname}{ptmri8t at 7pt}
\let\crnotice\mycrnotice%
\let\confname\myconfname%

\usepackage{microtype}

\makeatletter
\def\@copyrightspace{\relax}
\makeatother

\begin{document}
\addtolength{\parskip}{-0.5mm}
\setlength{\textfloatsep}{0cm}

\title{Leveraging Textual Features for Best Answer Prediction in Community-based Question Answering}

\numberofauthors{2}
\author{
\alignauthor
George Gkotsis, Carlos Pedrinaci, John Domingue\\
       \affaddr{Knowledge Media Institute}\\
       \affaddr{The Open University}\\
       \affaddr{Milton Keynes, UK}\\
       \email{firstname.lastname@open.ac.uk}
\alignauthor
Maria Liakata\\
       \affaddr{Dept. of Computer Science}\\
       \affaddr{University of Warwick}\\
       \affaddr{Coventry, UK}\\
       \email{m.liakata@warwick.ac.uk}
}

\maketitle

One of the intriguing problems in Community-based Question Answering (CQA) research is the automatic identification of the best answer, which is expected to benefit various stakeholders. First of all, since several answers are provided for each question, the readers of these websites will be able to process the candidate answers more efficiently and mitigate the ``information overload'' phenomenon. Secondly, a mechanism that identifies high quality answers will increase awareness within the community and will help to put more effort into questions that remain poorly answered. For instance, in StackOverflow\footnote{\url{http://stackoverflow.com/}}(SO) alone, as of September 2013, we found that approximately 33\% of the questions have yet to be marked as resolved (i.e., out of the 5 million, 1.7 million questions have no answer marked as ``accepted'').

Researchers in related fields have used lexical, syntactic, and discourse features to produce a predictive model of readers' judgments~\cite{pitler2008revisiting}. In several cases, the use of shallow features, i.e. features that do not employ semantic or syntactic parsing such as sentence length or word length, have been shown to be effective in assessing properties such as ease of reading or usefulness. However, with respect to CQA, research efforts towards the exploitation of shallow features report relatively low results. To improve the efficacy of their models, researchers refer to more contextual information, such as the \emph{score} of each answer, the \emph{comments} received or the \emph{reputation} of the user~\cite{anderson2012discovering}. However, these features may not be readily available since a) comments and scores introduce an inherent delay, and b) features based on reputation may not be applicable on a newly formed community or pose a threat to its development (i.e. preferential attachment) and result in the reinforcement of the pre-existing community hierarchy.

\begin{figure}
\centering
    \includegraphics[scale=.35]{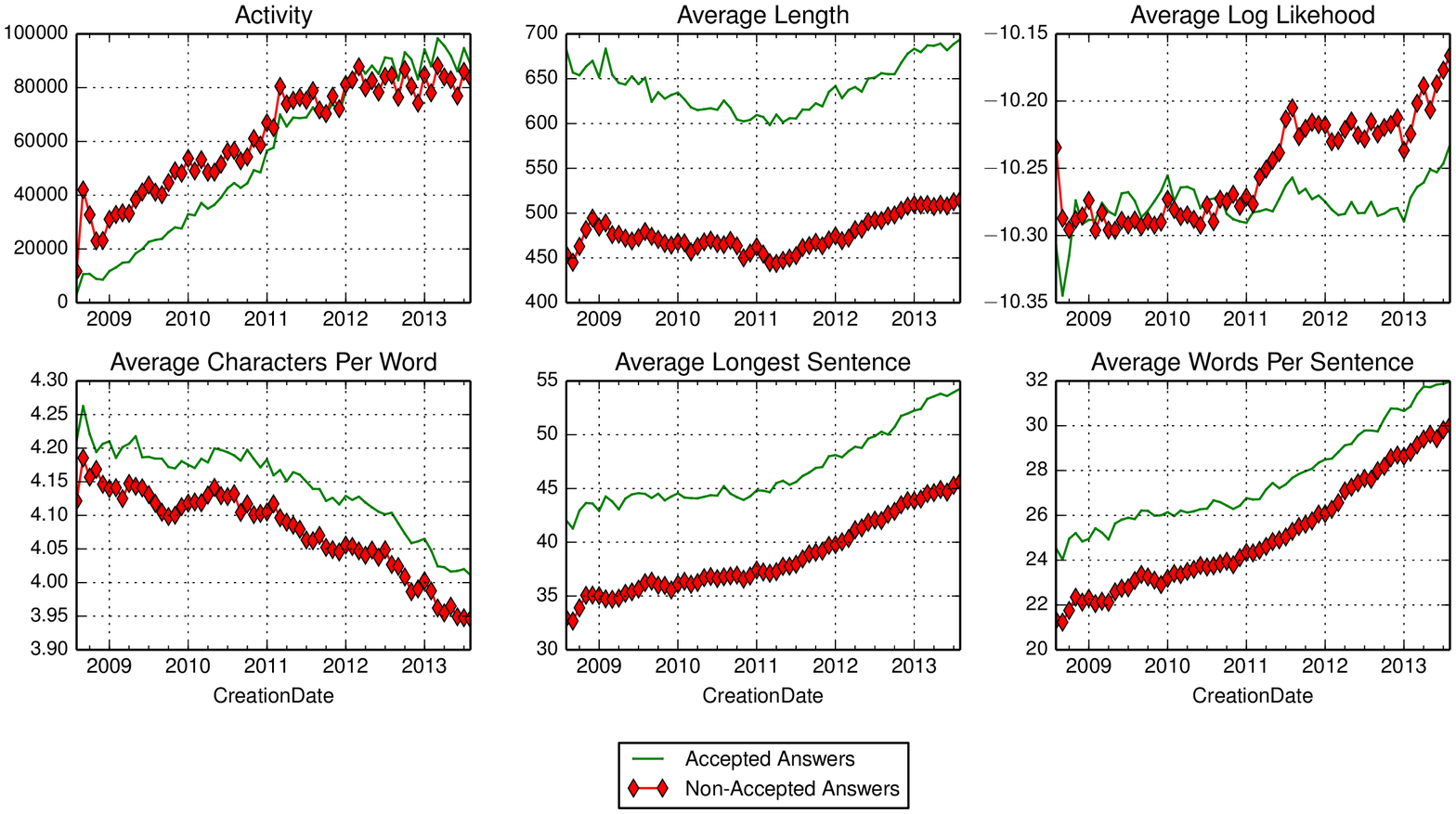} 
\caption{Activity and values of the linguistic features (y-axis) for the StackOverflow dataset over time~(x-axis). Top left sub-plot shows the number of answers posted every month. The remaining sub-plots show the average values  for the accepted and non-accepted answers.}
\label{so-linguistics}
\end{figure}

In our approach, we revisit the case of shallow linguistic features and use features found in~\cite{pitler2008revisiting}. Figure~\ref{so-linguistics} shows the average feature values for the accepted answers together with the non-accepted ones of SO using a one-month window time frame\footnote{Similar behaviour is identified for all StackExchange websites and is omitted due to space limitations.}. As seen from the figure, the linguistic features clearly differentiate the accepted from the non-accepted answers. More specifically, accepted answers tend to be longer, use a less common vocabulary, contain longer words, more words per sentence and the longest sentences are lengthier. Even though the above remarks look promising concerning best answer prediction, when training a binary classifier \emph{prediction} remains weak (58\% precision and 0.56 F-Measure on average for all StackExchange - SE - websites). A more thorough investigation towards the explanation of this poor performance leads us to identify two main issues. Firstly, as illustrated in Figure~\ref{so-linguistics}, the characteristics of language  evolve over time. Secondly, while a steady gap between the average values of accepted and non-accepted endures, a large inherent diversity of the posts persists together with a large variance. Finally, a cross-examination of absolute values between different SE websites has shown us that language characteristics differ significantly across SE websites. Since the results that we obtained for a classification based on shallow features are comparable to similar approaches (e.g.~\cite{tian2013towards}) these results will constitute our baseline for evaluating the proposed solution.

\vspace{-0.2cm}
\section{Feature discretisation}
\label{features}
\vspace{-0.1cm}
Our solution called \emph{discretisation} is presented in detail in~\cite{gkotsis2014} and asserts the adoption of a novel way of leveraging shallow features and overcome the above limitations. Intuitively, our approach is to treat the collection of answers for \emph{each question} as an \emph{information unit} which can improve the training process. Instead of treating each answer independently of the other answers it is competing with, our approach is to assess the value of the features of each answer \emph{in relation} to the corresponding features of its competitors. We introduce a new set of features that stem from the linguistic features used so far: instead of dealing with continuous values, these new features are the result of \emph{grouping}, \emph{sorting}, and \emph{discretisation}. 

We will present an example for the $Length$ feature. Let us consider the example of Table~\ref{linguistic-dis} where for one question there are two candidate answers (i.e., question with Id 5 having answers with Id 6 and 7). We have already shown previously (Figure~\ref{so-linguistics}) that the longer an answer is, the more likely it is to be accepted. In order to represent this preference, we group all answers by their corresponding questions (\emph{grouping}). For each group, we then sort the answers (\emph{sorting}) and assign a rank for each answer, starting from 1 and incrementing this rank by 1 (\emph{discretisation}). Sorting is done either in descending or ascending order, so as the lowest rank is assigned to the answers that are marked as accepted (in this example, we use the information that longer answers are more likely to be accepted, hence descending order is conducted). For the example of Table~\ref{linguistic-dis}, the answer with the longest $Length$ will receive $Length_{D}$ of value 1 (answer Id 6 with length 250) while the answer that comes second a value of 2 (answer Id 7 with length 200~-~note that we are representing the discretised form of each $feature$ as $feature_{D}$). The result of this process is the introduction of an equal number of linguistic features without the usage of any further information (apart from the association of a question and its corresponding answers\footnote{Note that other approaches typically omit this information.}).

\begin{table}
\caption{Example of feature discretisation for the case of $Length$, 5 submitted answers and 2 questions. Column Question Id refers to the question under which the answer is submitted.}
\begin{center}
{\scriptsize
\begin{tabular}{|l|l|l|l|}
\hline
Question Id & Answer Id & $Length$ & $Length_{D}$\\\hline\hline
 \multirow{3}{*}{1} & 2 & 200 & 2\\
 & 3 & 150 & 3\\
 & 4 & 250 & 1\\\hline
 \multirow{2}{*}{5} & 6 & 250 & 1\\
 & 7 & 200 & 2\\\hline
\end{tabular}
}
\end{center}
\label{linguistic-dis}
\end{table}%

\vspace{-0.2cm}
\section{Evaluation}
\vspace{-0.1cm}
Table~\ref{results-features} presents the results when using different sets of features and 10-fold validation. The table contains the average values for 21 SE websites (including SO) as the output of different evaluations on 4 million questions and more than 8 million answers. Initially, we use the absolute values of textual features with low results (58\% precision, Case~1). The second and third Cases both utilise the discretised features, while the third is additionally using the \emph{other} set of features (i.e. AnswerCount and CreationDate). Cases~2 and 3 constitute our proposed prediction method. Furthermore Case~4 refers to a ``traditional'' approach that relies in plain linguistics \emph{and} user-reputation ratings. We can see that while a whole new set of features is added into the dataset, the performance of classification remains lower than Case~3, which is linguistics-based. Case~5 keeps the user ratings in addition to incorporating all features of Case~3. Hence, classification accuracy is the highest compared to all previous classifications, but almost identical to Case~3 which is strictly based on content and discretisation (higher F-Measure 0.77 vs. 0.76, higher AUC 0.88 vs. 0.87). Finally, Case~6 uses all previous features, including the \emph{answer ratings}. This set of features uses all features but most importantly user-entered scores and manages to outperform all of the previous cases. Case~6 shows that the information contained within answer ratings is independent  -- to a certain extent -- of the information found in previous features.

\begin{table}
\caption{Results for best answer prediction using different sets of features (Cases~1 to 6) for all SE websites. Columns show macro average precision (P), recall (R), F-Measure (FM) and Area-Under-Curve (AUC) for all 21 SE websites using 10-fold validation.}
\begin{center}
{\scriptsize
\begin{tabular}{|l|l|l|l|l|l|}
\hline
No. & Features Used & P & R & FM & AUC\\\hline\hline
1 & Linguistic & 0.58 & 0.60 & 0.56 & 0.60\\\hline
2 & {\parbox{2.5cm}{\vspace{.1\baselineskip}Linguistic \&\\ Discretisation\vspace{.1\baselineskip}}} & \cellcolor{gray!25}0.81 & \cellcolor{gray!25}0.70 & \cellcolor{gray!25}0.74 & \cellcolor{gray!25}0.84\\\hline
3 & {\parbox{2.5cm}{\vspace{.1\baselineskip}Linguistic \&\\ Discretisation \&\\ Other\vspace{.1\baselineskip}}}& \cellcolor{gray!25}0.84 & \cellcolor{gray!25}0.70 & \cellcolor{gray!25}0.76 & \cellcolor{gray!25}0.87 \\\hline
4 & {\parbox{2.5cm}{\vspace{.1\baselineskip}Linguistic \& Other \& User Rating \\ (no discretisation)\vspace{.1\baselineskip}}} & 0.82 & 0.69 & 0.75 & 0.86\\\hline 
5 & {\parbox{2.5cm}{\vspace{.1\baselineskip}Linguistic \& Other \& User Rating \\(with discretisation)\vspace{.1\baselineskip}}}& 0.82 & 0.72 & 0.77 & 0.88\\\hline 
6 & {\parbox{2.5cm}{\vspace{.1\baselineskip}All features\\ (Answer and User Rating with discretisation)\vspace{.1\baselineskip}}} & 0.88 & 0.85 & 0.86 & 0.94 \\\hline 
\end{tabular}
}
\end{center}
\label{results-features}
\end{table}%

In summary, results in Table~\ref{results-features} show that the discretisation of linguistic features manages to outperform significantly the classifier based on linguistic features only. Moreover, we can also see that user rating features such as reputation do not improve our classification, a sign that discretisation is a process that extracts very useful information and delivers very strong results.

The whole approach described here has been implemented and is offered for free as web browser plugin and a web service (\url{https://acqua.kmi.open.ac.uk}). In the future, we intend to explore the applicability of our methodology elsewhere and investigate further the effect of textual quality on answer selection and impact in online fora and social media.

%
%
%
%
%
%
%

\section*{Acknowledgments}
This work was supported by the CARRE (No.611140) and COMPOSE (No.317862) projects funded by the European Commission Framework Programme~7.

\bibliographystyle{abbrv}
\small
\bibliography{sigproc} 

\begin{thebibliography}{1}

\bibitem{anderson2012discovering}
A.~Anderson, D.~Huttenlocher, J.~Kleinberg, and J.~Leskovec.
\newblock Discovering value from community activity on focused question
  answering sites: a case study of stack overflow.
\newblock In {\em Proceedings of the 18th ACM SIGKDD international conference
  on Knowledge discovery and data mining}, pages 850--858. ACM, 2012.

\bibitem{gkotsis2014}
G.~Gkotsis, K.~Stepanyan, C.~Pedrinaci, J.~Domingue, and M.~Liakata.
\newblock It's all in the content: State of the art best answer prediction
  based on discretisation of shallow linguistic features.
\newblock In {\em Proceedings of the 2014 ACM Conference on Web Science},
  WebSci '14, pages 202--210, New York, NY, USA, 2014. ACM.

\bibitem{pitler2008revisiting}
E.~Pitler and A.~Nenkova.
\newblock Revisiting readability: A unified framework for predicting text
  quality.
\newblock In {\em Proceedings of the Conference on Empirical Methods in Natural
  Language Processing}, pages 186--195. Association for Computational
  Linguistics, 2008.

\bibitem{tian2013towards}
Q.~Tian, P.~Zhang, and B.~Li.
\newblock Towards predicting the best answers in community-based
  question-answering services.
\newblock In {\em Seventh International AAAI Conference on Weblogs and Social
  Media}, 2013.

\end{thebibliography}

\end{document}